\documentclass[journal,twoside,confmode]{ieeecolor}

\usepackage{tmi}

\usepackage{cite}
\usepackage{amsmath,amssymb,amsfonts}
\usepackage{algorithmic}
\usepackage{graphicx}
\usepackage{textcomp}
\usepackage{hyperref}
\usepackage{doi}
\usepackage[capitalise]{cleveref}
\Crefname{section}{Sec.}{Secs.}
\usepackage{multirow}
\usepackage{underscore}
\usepackage{graphicx} 
\usepackage{booktabs}
\usepackage{colortbl}
\usepackage{makecell}
\usepackage{subcaption} 
\usepackage{float}

\newcommand{\cint}[2]{
  \makecell[tc]{
    #1\\[-2pt] 
    {\scriptsize\textcolor[gray]{0.8}{[#2]}}%
  }
}

\makeatletter
\def\@IEEEpubidpullup{0pt}
\def\IEEEpubid#1{}
\makeatother
\begin{document}
\title{Multi-View Stenosis Classification Leveraging Transformer-Based Multiple-Instance Learning Using Real-World Clinical Data}
\author{N. Cenikj, Ö. Turgut, A. Müller, A. Steger, J. Kehrer, M. Brugger, D. Rueckert, \IEEEmembership{Fellow, IEEE}, E. Martens, and P. Müller
\thanks{This work has been submitted to the IEEE for possible publication. Copyright may be transferred without notice, after which this version may no longer be accessible.}
\thanks{Nikola Cenikj, Özgün Turgut, Daniel Rueckert, and Philip  Müller are with Chair for AI in Healthcare and Medicine, Technical University of Munich (TUM) and TUM University Hospital, Munich, Germany. (e-mail: nikola.cenikj@tum.de; oezguen.turgut@tum.de; daniel.rueckert@tum.de; philip.j.mueller@tum.de)}
\thanks{Daniel Rueckert is also with the Department of Computing, Imperial College London, UK, and Munich Center for Machine Learning (MCML), Munich, Germany.}
\thanks{Nikola Cenikj, Alexander Müller, Alexander Steger, Jan Kehrer, Marcus Brugger, and Eimo Martens are with the Department of Internal Medicine, TUM University Hospital, Munich, Germany. (e-mail: alexander.mueller@mri.tum.de; alexander.steger@tum.de; jan.koehlen@gmx.de; marcus.brugger@mri.tum.de; eimo.martens@mri.tum.de)}}

\maketitle
\begin{abstract}
Coronary artery stenosis is a leading cause of cardiovascular disease, diagnosed by analyzing the coronary arteries from multiple angiography views.
Although numerous deep-learning models have been proposed for stenosis detection from a single angiography view, their performance heavily relies on expensive view-level annotations, which are often not readily available in hospital systems. Moreover, these models fail to capture the temporal dynamics and dependencies among multiple views, which are crucial
for clinical diagnosis. To address this, we propose SegmentMIL, a transformer-based multi-view multiple-instance learning framework for patient-level stenosis classification. Trained on a real-world clinical dataset, using patient-level supervision and without any view-level annotations, SegmentMIL jointly predicts the presence of stenosis and localizes the affected anatomical region, distinguishing between the right and left coronary arteries and their respective segments. 
SegmentMIL obtains high performance on internal and external evaluations and outperforms both view-level models and classical MIL baselines, underscoring its potential as a clinically viable and scalable solution for coronary stenosis diagnosis. Our code is available at \href{https://github.com/NikolaCenic/mil-stenosis}{https://github.com/NikolaCenic/mil-stenosis}.


\end{abstract}

\section{Introduction}
\IEEEPARstart{D}{espite} significant advancements in the diagnosis and treatment of cardiac
diseases, \emph{coronary artery stenosis} is one of the major causes of impaired cardiac function and reduced
patient life expectancy. 
\begin{figure}[!t]
  \centering
     \includegraphics[width=.94\linewidth]{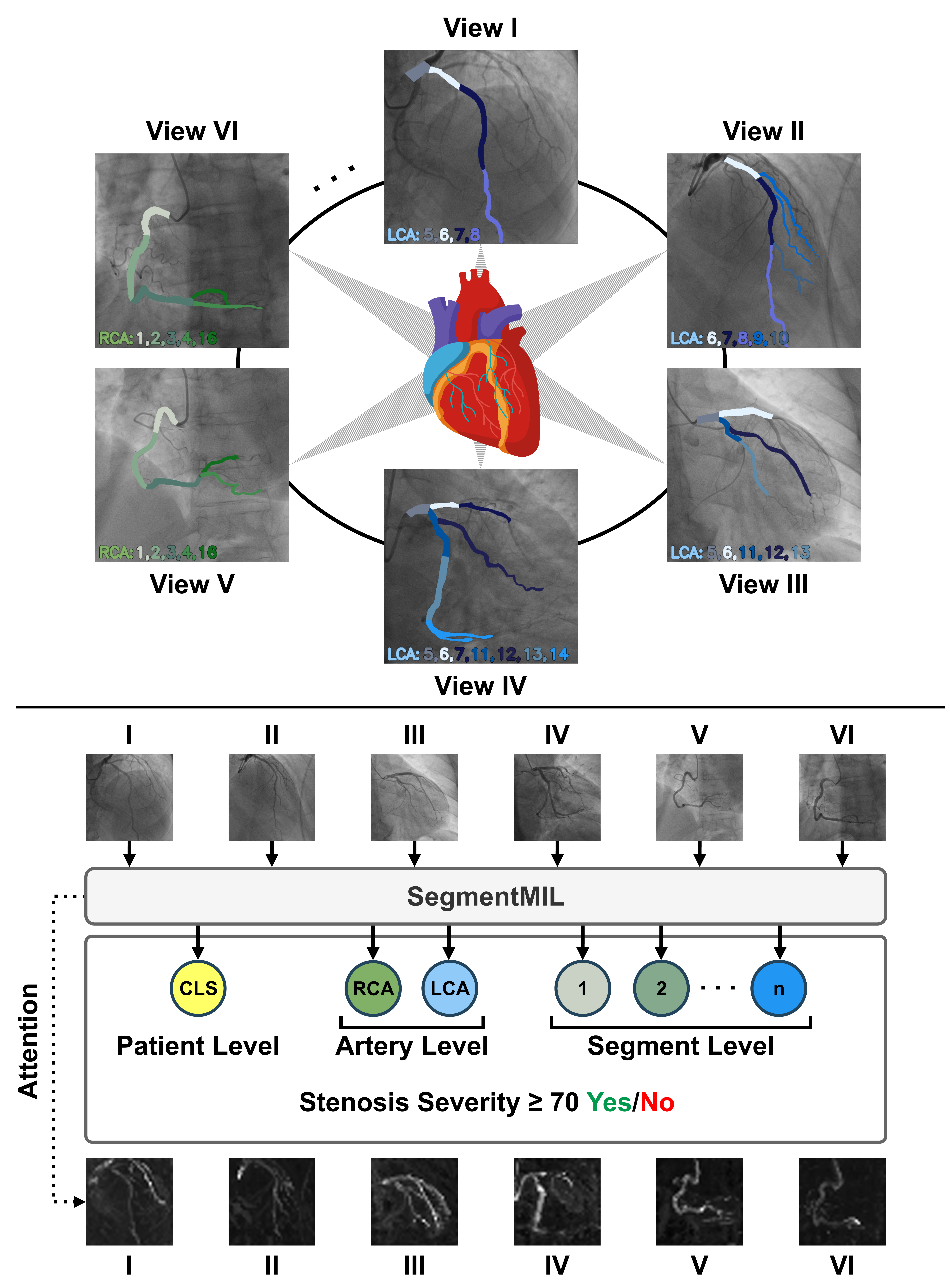}
     \caption{Overview of our approach: Multiple angiographic views from a single patient capture the coronary arteries from various angles, providing information on different segments of the coronary arteries, highlighted in green for the right (RCA) and blue for the left (LCA) coronary artery. Since a single view includes multiple segments and each segment appears in multiple views, patient-level stenosis diagnosis requires an integrated analysis of all views. To address this, we propose \emph{SegmentMIL}, a multi-view transformer-based stenosis classification model capable of predicting patient-, artery-, and segment-level stenosis. Furthermore, by leveraging the transformer's attention maps, we derive zero-shot artery segmentation masks, providing interpretable visual explanations of the model’s decision process. }\label{fig:main}
\end{figure}
Coronary artery stenosis is characterized by the narrowing of the coronary arteries, which restricts blood flow to the heart muscle, causing chest pain, shortness of breath, and, in severe cases, myocardial infarction. 
Studies show a five-year survival rate of 73\% for stenosis patients~\cite{survival_rate}, with rates varying by severity: 92\% for single-vessel, 65\% for double-vessel, and 55\% for triple-vessel stenosis, emphasizing the need for timely diagnosis. The diagnosis involves cardiac catheterization, where a contrast agent is injected into the right (RCA) and left (LCA) coronary arteries, and is tracked using temporal sequences of \emph{angiographic X-rays}. 
Given the criticality of timely diagnosis, automated systems capable of operating continuously could provide rapid preliminary assessments and aid as complementary tools for reliable clinical decision-making.

Even though stenosis diagnosis has already been subject of research in the deep-learning domain\cite{stenosis_detection_extra_1,stenosis_detection_extra_2,stenosis_detection_extra_spatio_temporal,stenosis_image_and_patient_level_cls,rca_stenosis}
 current methods cannot be trained using existing hospital annotations and thus require diverse manually annotated datasets. In addition, these models focus on stenosis detection from a single frame from a single view, ignoring the temporal dynamics and the dependencies between different views, which contain critical information about the contrast agent flow required for accurate diagnosis. Furthermore, current models do not contain fine-grained stenosis diagnoses on artery- or segment-level, which can provide useful information for clinical decision making.

To address these limitations, we propose \emph{SegmentMIL}, a multi-view stenosis classification model leveraging transformer-based multiple-instance learning (MIL). As shown in \cref{fig:main}, our model predicts patient-, artery-, and segment-level stenosis from the full set of available multi-view angiographic X-ray sequences per patient.
It is trained on raw clinical data with targets derived directly from the hospital data system, without the need for any additional manual labeling. As hospital system targets were used for real-world clinical decision-making, they serve as highly reliable annotations.

Our contributions are as follows:
\begin{enumerate}
\item We propose \emph{SegmentMIL}, 
a transformer-based stenosis
classification model predicting patient-, artery-, and segment-level stenosis from multi-view coronary angiographies.
\item We enable the analysis of temporal dynamics by jointly processing multiple frames per view, located around a detected key frame, not supported by other methods.
\item We thoroughly analyze the prediction quality of our model on both an internal test set as well as on a public test set, comparing it to common MIL approaches and to view-level trained baselines. 
Our SegmentMIL model outperforms the baselines by large margins, achieving especially high quality when using multiple frames.
\item We provide extensive ablation studies to study the impact of the proposed design decisions.

\end{enumerate}

\section{Related Work}
\subsection{Stenosis Classification}
The field of deep-learning-based stenosis diagnosis has significantly advanced with the development of image-based models, with research targeting classification, segmentation, and detection tasks. 
A pivotal milestone in this field is the ARCADE Challenge~\cite{ARCADE}, which introduced a benchmark of coronary angiography images for artery and stenosis segmentations. The leading stenosis segmentation model~\cite{arcade_1} on ARCADE achieves an F1 of 0.57. The work in~\cite{stenosis_detection_overview} evaluates multiple widely used detector networks, reporting F1 of 0.96 when using a Faster-RCNN~\cite{fasterrcnn} model. 
Other works also address stenosis detection~\cite{stenosis_detection_extra_1,stenosis_detection_extra_2,stenosis_detection_extra_spatio_temporal}, with \cite{stenosis_detection_extra_spatio_temporal} being the only one to exploit intra-view temporal information.
In view-level stenosis classification, \cite{rca_stenosis} achieves an AUC of 0.925, but only considering the RCA. 
The CADICA dataset~\cite{cadica} advanced the research in this field by providing frame-level severity and localization 
labels. A ResNet-50 model~\cite{resnet} trained on this dataset
achieved F1 scores of 0.83 (RCA) and 0.81 (LCA).
The work in~\cite{stenosis_3_view_quantification} quantifies stenosis using one main and one support view. Although it is the only method incorporating multiple views, it is constrained to a two-view setup.
To the best of our knowledge, \cite{stenosis_image_and_patient_level_cls} is the only study that, beyond view-level, also reports artery- and patient-level performances. For the LCA, they train four separate models on views from four specific angulations of a patient, with predictions aggregated using max-pooling. For the RCA, they use a single model applied to three views, with aggregation performed in the same manner. Similarly,  patient-level predictions are derived by max pooling over the RCA and LCA outputs. In such a controlled setup, they report AUCs of 0.89 and 0.84 for the RCA and LCA, and 0.86 at the patient level. However, no approaches have been specifically trained for patient-level diagnosis. In practice, cardiologists diagnose stenosis by examining the contrast agent flow in many different angulations, emphasizing the need for patient-level models that align with current clinical workflows.
\subsection{Multiple-instance Learning}
Multiple-instance learning (MIL) is a weakly supervised learning framework, where samples are organized into bags, with a single label assigned to each bag. Existing MIL approaches can be broadly grouped based on their bag-level aggregation strategy, distinguishing between instance- and embedding-level methods. 
Instance-level methods perform predictions at the instance level and aggregate them using pooling operations~\cite{mil_instance_level,max_mil,max_vs_noisy_or,mil_instance_level2}, which limits their ability to capture relationships between instances.
In contrast, embedding-level MIL methods aggregate instances in a latent space, forming bag-level feature representations. 
This is typically done using graph-based~\cite{heat, bazargani2023multiscalerelationalgraphconvolutional, lymph_node_graph_based_mil} or attention-based methods~\cite{mil_base, attention, dual_stream_embedding_mil, leukemia-cls-attention-mil}. A key advantage of attention-based approaches is that the instance-level attention scores show the contribution of each instance to the final prediction~\cite{rethinking_attention_based_mil, pamil, xmil, clam}.
In the medical domain, the MIL paradigm has been widely used in histopathology, where gigapixel whole slide images cannot be processed in a single pass and are instead divided into multiple patches, treated as a bag of samples. 
Such an approach has been used for cancer detection\cite{retmil}, cancer survival prediction~\cite{cancer_survival_probability, lung_cancer_survival_probability,MMsurv}, as well as pathology report generation~\cite{pathology_report_generation}. 
However, despite its success in weakly supervised medical imaging, MIL has not been applied in the angiography domain, even though multi-view angiography data fully aligns with the MIL paradigm.
\section{Materials and Methods}
\subsection{Datasets}\label{sec:dataset}
\begin{figure}[t!]
     \centering
        \includegraphics[width=\linewidth]{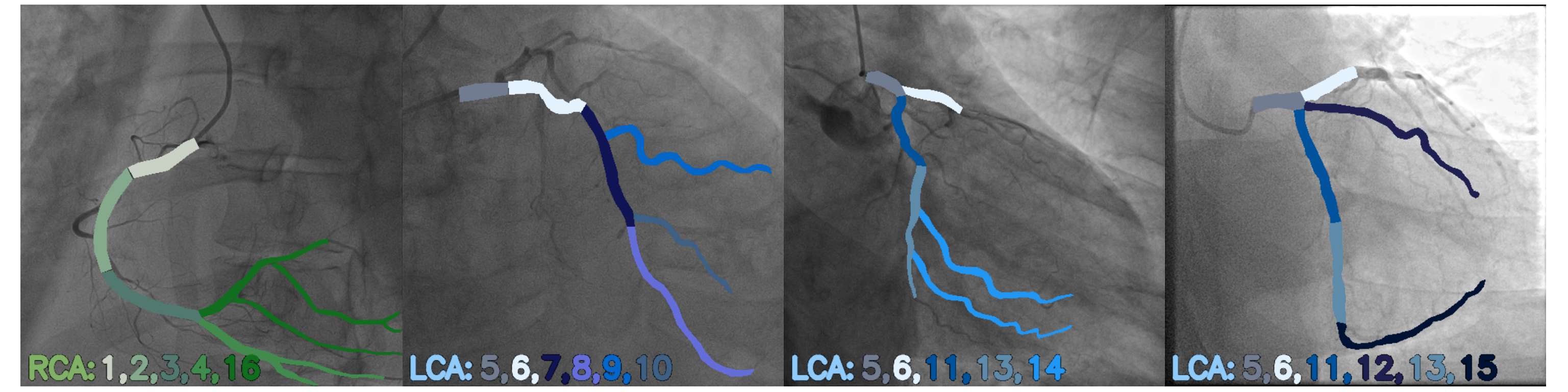}%
        \caption{Example frames from angiography views with the coronary arteries highlighted. The coronary artery system consists of two main branches: the right (RCA, marked in green) and the left coronary artery (LCA, marked in blue). Based on the Syntax Score Methodology\cite{sintax_score}, RCA and LCA are divided into 16 segments. Segments 1, 2, 3, 4, and 16 correspond to the RCA, and the remaining belong to the LCA. }
        \label{fig:cad_seg}
    \end{figure}
\subsubsection{Clinical Dataset}
\label{section:clinical_dataset}
We use a clinical dataset containing 17,741 angiography views of 2,003 patients (median age of 71, and 70\% male), treated at the TUM Klinikum Rechts der Isar, Munich, Germany. The angiography views we consider are acquired as video sequences during the first 15 minutes of a single cardiac catheterization. Each patient is labeled with segment-level stenosis severity for the 16 segments of the coronary arteries, defined by the Syntax Score\cite{sintax_score} methodology. Example frames highlighting the arteries and corresponding segments are shown in \cref{fig:cad_seg}. Since the stenosis severities cannot be mapped to continuous targets for a regression task, they are discretized into seven severity categories, similar to the CAD-RADS classification~\cite{cad-rads}. The used severity categories are: $\geq$0, $\geq$20, $\geq$50, $\geq$70, $\geq$90, 99, and 100. The labels are highly reliable as they have been used for real-world decision making, specifying only the severity category without any information about the view or location of the stenosis. 
As the segments differ in size, and the smaller segments are difficult to annotate, in this study, we focus on the larger segments, to which we refer as major,  $S_{\mathrm{major}} =  {\{1,2,3,5, 6, 7, 11, 13\}}$, selected in consultation with experts. Based on the artery they belong to (RCA or LCA), the  $S_{\mathrm{major}}$ segments can be further split into 
 $S_{\mathrm{RCA}} = \{1, 2, 3\}$ and $S_{\mathrm{LCA}} = \{5, 6, 7, 11, 13\}$.
Given the severity categories of the $S_{major}$ segments of a patient, we obtain artery and patient-level severities defined as the maximal severity in the segments belonging to the artery or patient. The severities are binarized with a threshold of 70\%, indicating highly relevant and severe stenosis. We split the dataset at the patient-level into train, validation, and test sets such that the validation and test sets contain 200 patients each and are perfectly balanced with respect to patient-level stenosis. The remaining patients, of which 694 exhibit severe stenosis, form the train set.
Even though the number of views per patient is spread over a wider range, the patients having too few or too many views are rare, and using them for evaluation could infer a bias (see \cref{sec:bias_from_view_cnt}). To ensure a comprehensive and reliable evaluation, the test set is selected to contain between 8 and 12 views per patient.

 We further utilize an additional test set from the same hospital, comprising 100 patients and 760 views, annotated by cardiologists for the exact location, segment, and severity of stenosis. As this dataset includes view-level annotations, we use it for view-level evaluation and refer to it as the view-level internal test set. After binarization, 133 of the 760 annotated views exhibit severe stenosis ($\ge$70\%).

\begin{figure}[!t]
    \centering

    \begin{subfigure}[b]{0.49\linewidth}
     \centering
        \includegraphics[height=3.5cm,width=\linewidth]{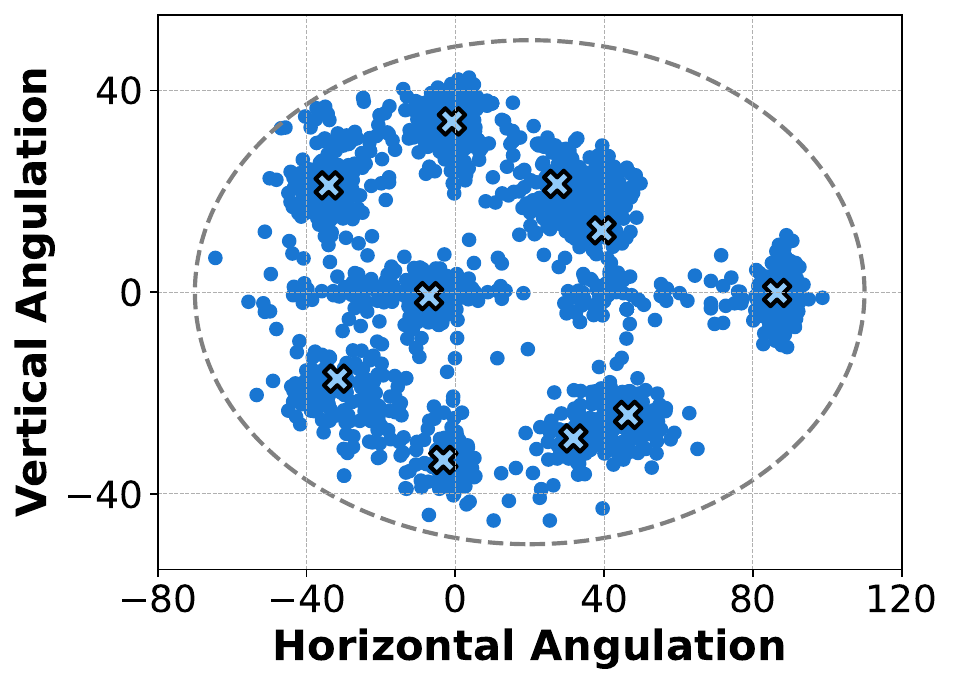}%
        \caption{Patient-Level  Test Set.}
        \label{subfig:ang_a}
    \end{subfigure}
    \begin{subfigure}[b]{0.49\linewidth}
     \centering
        \includegraphics[height=3.5cm,width=\linewidth]{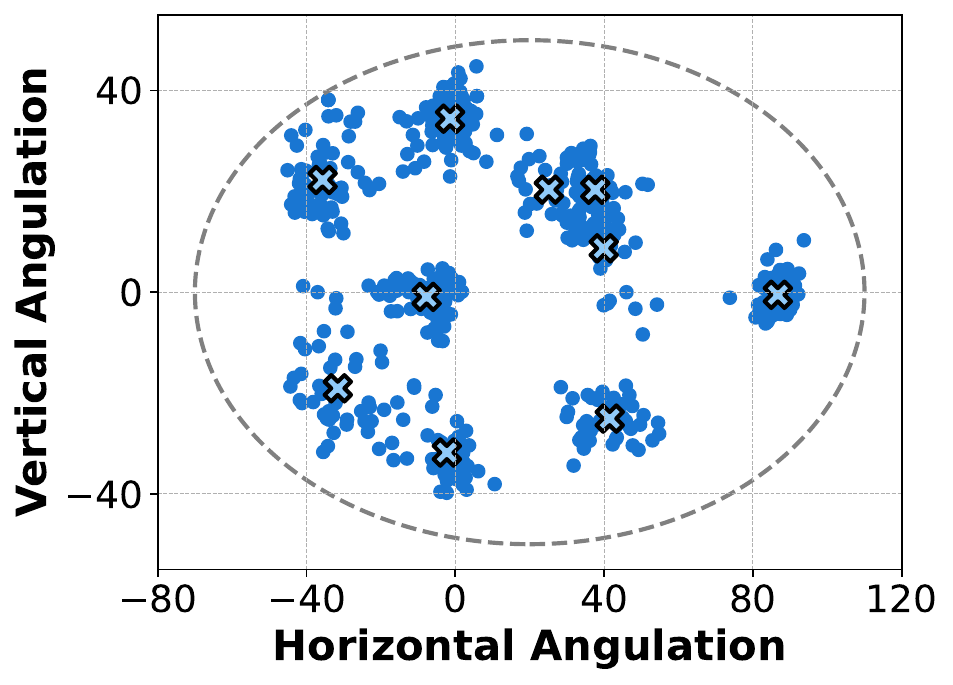}%
        \caption{View-Level Test Set.}
        \label{subfig:ang_b}
    \end{subfigure}

    \caption{Distribution of the horizontal and vertical angulations across views.
The angulations correspond to the positioning of the C-arm of the X-ray device used to image the coronary angiography. The angulation clusters are highlighted by the K-means centroids (denoted by $\pmb{\times}$), showing the 
clinical practice, where acquisitions are performed from standardized angulations for visualizing specific coronary arteries and segments.}
    \label{fig:data_angulation}
\end{figure}

The DICOM header of the views from the clinical data contains information about the horizontal and vertical angulations of the view, which correspond to the location of the C-arm of the X-ray device used for angiography capturing.
In \cref{fig:data_angulation} we show the angulations of all views across the evaluation datasets. To highlight the present clusters, we plot the K-means centroids. The consistency between the clusters in the two sets reflects clinical practice, in which angiography acquisitions are performed from standardized angulations optimized for visualizing specific coronary arteries and segments.

\subsubsection{CADICA Dataset}\label{section:cadica_dataset}
 To extend the evaluation scope, we employ CADICA~\cite{cadica}, a publicly available dataset collected from a hospital in Malaga, Spain. The dataset includes view-level annotations specifying the location and severity category of stenosis for 382 views of 42 patients. Among these, 122 views exhibit severe stenosis ($\ge$70\%), which corresponds to 28 patients. Unlike our internal dataset, CADICA does not provide the angulations of the views. Since the annotations lack segment-level details, this dataset is used exclusively as an external test set for patient- and view-level evaluation.

\begin{figure}[t!]
    \centering

    \begin{subfigure}[b]{0.49\linewidth}
     \centering
        \includegraphics[width=\linewidth]{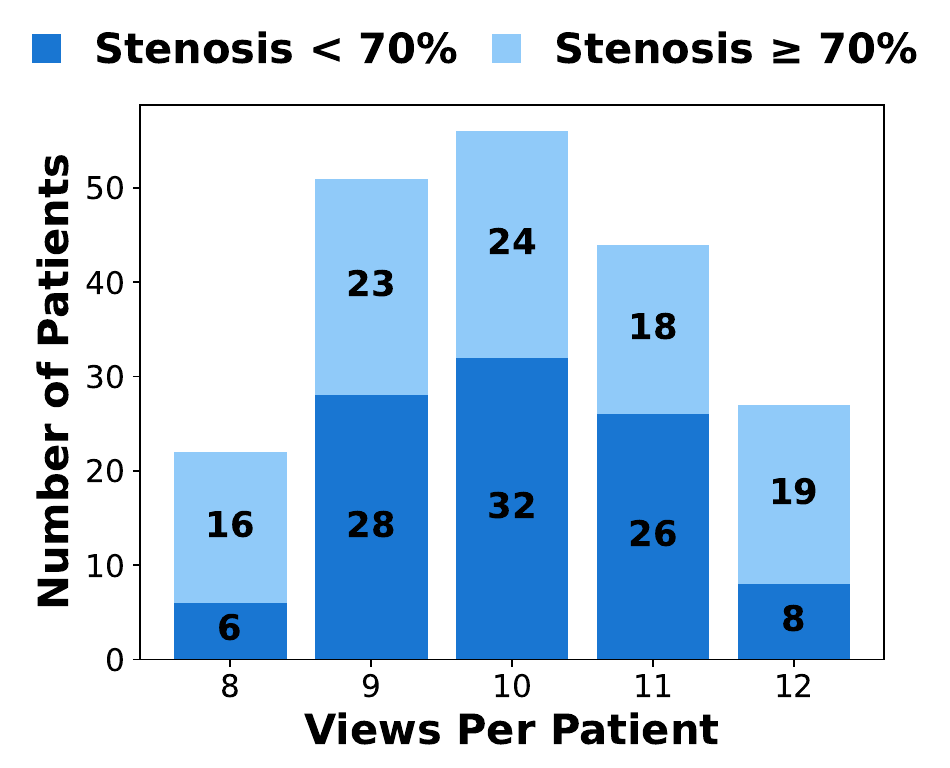}%
        \caption{Internal Test Set.}
        \label{fig:data:a}
    \end{subfigure}
    \begin{subfigure}[b]{0.49\linewidth}
     \centering
         \includegraphics[width=\linewidth]{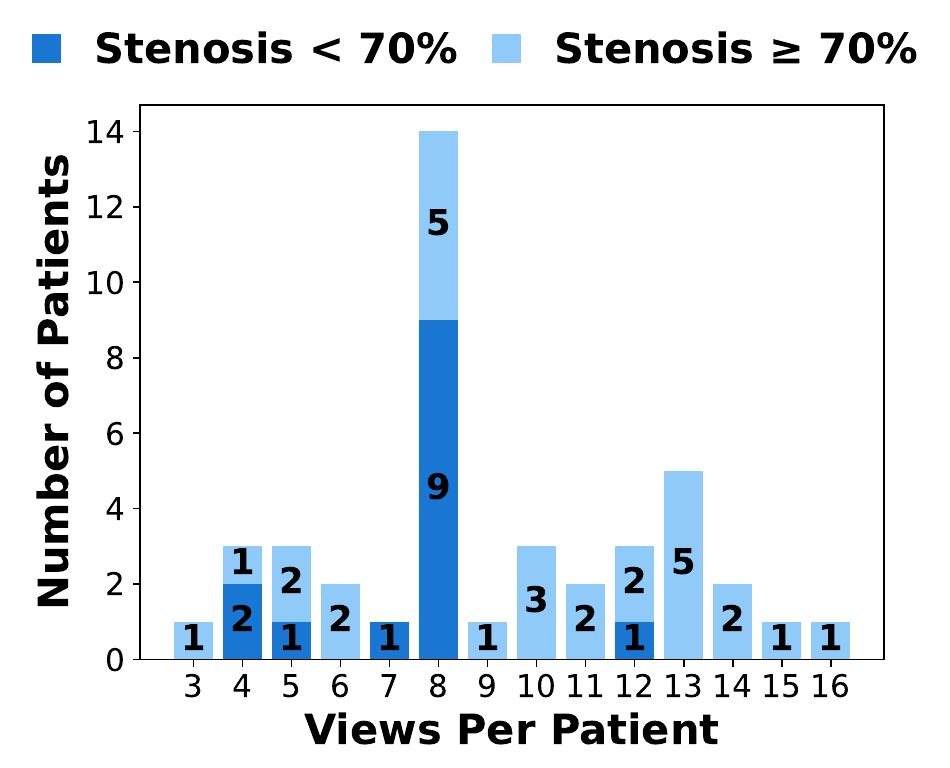}%
        \caption{CADICA Test Set.}
        \label{fig:data:b}
    \end{subfigure}
    \caption{ Comparison of stenosis distribution for patients with different number of views. The internal test set (a) is selected to have a tight range of views-per-patient, in order to obtain a more representative evaluation. The CADICA test set distribution (b) hints that stenosis is more common among patients with more views. To evaluate such bias, we trained an XGBoost classifier to predict stenosis based only on the number of views and showed that this bias does not have a significant influence on evaluation performance.}
    \label{fig:data_distribution}
\end{figure}

 \subsubsection{Assessing Bias from Patient View Counts}\label{sec:bias_from_view_cnt}
In \cref{fig:data_distribution} we show the stenosis distribution for patients with
different numbers of views across the internal (\cref{fig:data:a}), and CADICA (\cref{fig:data:b}) test sets.
We assess whether the number of views and their angulations introduce a bias. Therefore, we train an XGBoost\cite{xgboost} classifier on the train set to predict stenosis based on (i) only the number of views, or (ii) based on the angulations of the presented views.
In (i), we achieve an AUC of 0.594 and 0.614 on the internal and CADICA test sets, and in (ii), an AUC of 0.572 on the internal test set.  
Despite biases being observed in \cref{fig:data:b} (e.g. higher presence of positive cases for patients with more than 9 views), these results show that the number of views and angulations alone are not enough to reliably predict the presence of stenosis, and thus, we do not expect this to significantly skew evaluations.

\subsection{Method}

\label{method}
\subsubsection{Overview}
\begin{figure*}[t!]
  \centering
     \includegraphics[width=\linewidth]{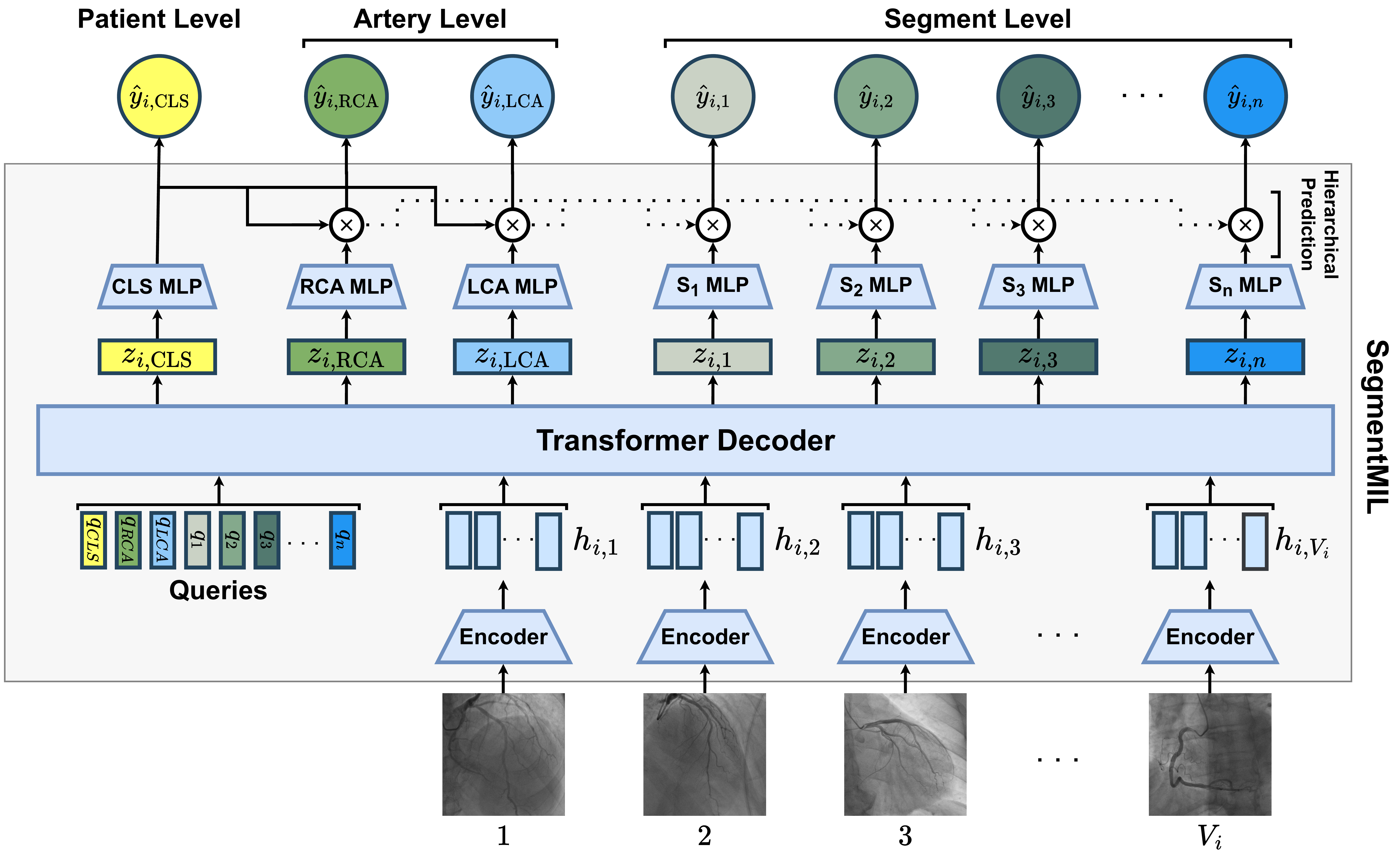}
      \caption{Architecture of the SegmentMIL model. Given the different views of a single patient, each view is encoded using a shared encoder. The encodings are then fed to a transformer decoder layer that uses learned queries to produce individual encodings for the different levels. Those encodings are further fed to unique classification heads, and the predicted probabilities are hierarchically merged, resulting in patient, artery, and segment-level stenosis predictions.}\label{Fig:methods}

  \end{figure*}

We propose \emph{SegmentMIL}, a transformer-based model designed for patient-, artery-, and segment-level stenosis classification based on multiple angiography views. An overview of the architecture of our SegmentMIL is shown in \cref{Fig:methods}. 
We first extract one key frame for each view (\cref{section:key_frame_algorithm}). Next, we encode each key frame individually using a shared ViT encoder (\cref{section:encoder}).  
 We use a transformer decoder\cite{attention} (\cref{section:decoder}) to aggregate the view-level embeddings into patient-, artery-, and segment-level representations, which are then fed to unique classification heads, predicting the presence of stenosis at the different levels (\cref{section:mlp_and_hierarchical}).

\subsubsection{Key Frame Detection}
\label{section:key_frame_algorithm}

\begin{figure}[t!]
     \centering
        \includegraphics[width=0.7\linewidth]{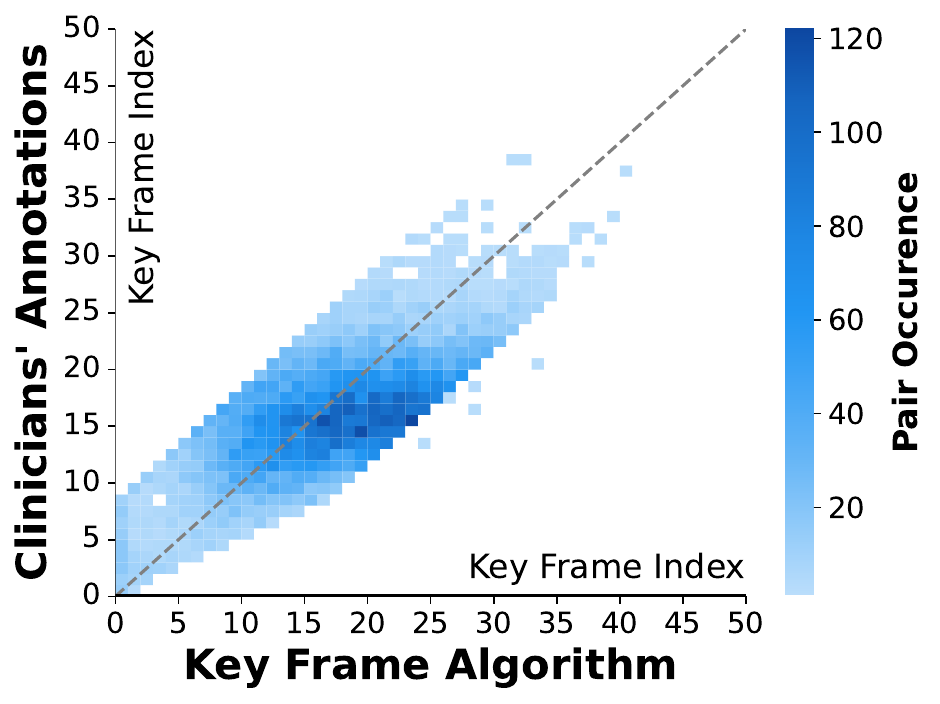}
        \caption{Comparison of the performance of our key frame detection algorithm ($x$-axis) against clinicians' annotations ($y$-axis). The absolute mean difference between the two is 3.77 frames, which corresponds to 0.53 seconds, given the frame rate of the angiography videos (7 frames per second).
        }
        \label{fig:key_frame}
    \end{figure}

Since the angiographic views contain temporal information and our initial focus is on image-based encoders, we developed a key frame detection algorithm, where a key frame is defined as the frame exhibiting the highest visible contrast agent within a view. Although the DICOM metadata of some of the views from the internal data contains clinicians' key-frame annotations, we aimed to create an independent system that does not rely on such information. 
Given a view, we first 
obtain artery segmentation masks for each frame using \emph{ArterySeg}, an artery segmentation model trained for this study using the ARCADE artery segmentation dataset~\cite{ARCADE}. We then rank the frames based on the surviving pixels in the segmentation masks, and as a key frame, select the one with the most surviving pixels.
We evaluated the correctness of our algorithm by comparing it to the expert annotations (\cref{fig:key_frame}). Our algorithm achieves an absolute mean difference of 3.77 frames, which, given the 7 frames per second rate of the internal data, corresponds to 0.53 seconds. 


\subsubsection{Encoder}
\label{section:encoder}
Consider patient $i$ with $V_{i}$ distinct angiographic views where each view is a $T$ frames $H \times W$-resolution video. For each view, we only select a single frame using our key frame detection algorithm (\cref{section:key_frame_algorithm}).
As encoder we use a ViT-S with a patch size of 14, initialized with  DinoV2\cite{dinov2}, shared across all frames. The encoder encodes each frame into patch embeddings  $\mathbf{h}_{i, v} \in \mathbb{R}^{N\times D}$, where $N = \frac{H}{14}\times \frac{W}{14}$ is the number of patches of a frame, and $D$ is the embedding size. 
\subsubsection{Transformer Decoder}
\label{section:decoder}
The transformer decoder aggregates the patch embeddings $h_{i,v}$ of each view, and obtains patient-level representations.
Given the patch embeddings of the $V_i$ views, it first adds spatial positional encodings. We use 2D fixed sinusoidal positional encodings, assigned based on the horizontal and vertical angulation of the view, that split the angulation plane into a $16\times 16$ grid.
The $V_i$ view patch embeddings are then fed as key and value tokens to the transformer decoder. 
As queries we use 11 learned tokens $\mathbf{q}_s \in \mathbb{R}^D$, 
one query for the patient (CLS), two for the RCA and LCA, and 8 for segment-level classification of the $S_{major}$ segments. The attention mechanisms within the transformer decoder aggregates the $V_i$ view embeddings $\mathbf{h}_{i, v}$ and the 11 queries $\mathbf{q}_s$ into 11 feature vectors $\mathbf{z}_{i, s} \in \mathbb{R}^{D}$, again one for patient, two for the main arteries and 8 for individual segments, capturing the global context for each of the three levels.
\subsubsection{MLP and Hierarchical Prediction}
\label{section:mlp_and_hierarchical}
The obtained feature representations $\mathbf{z}_{i, s}$ are subsequently passed through MLP classification heads (individually learned for each $s$) followed by sigmoid, yielding class probabilities $\tilde{p}_{i, s}$ for the patient-, artery-, and segment-level stenosis. 
To capture the hierarchical dependencies of coronary stenosis, where the presence of a lesion in any segment implies stenosis at the corresponding artery and patient levels, we introduce a two-level hierarchical prediction scheme. Inspired by the hierarchical softmax formulation~\cite{hierarchical_softmax}, artery-level predictions $\hat{y}_{i, s}, s \in \{{\mathrm{RCA}, \mathrm{LCA}\}}$, are obtained by multiplying the outputs of the artery-specific classifiers with the patient-level prediction. Similarly, segment-level predictions $\hat{y}_{i, s},s \in S_{\mathrm{major}}$, are computed as the product of the segment classifier output and the artery-level prediction of the artery to which the segment belongs, i.e. 
\begin{align}
    \hat{y}_{i, s} = \begin{cases}
        \tilde{p}_{i, \mathrm{s}} & \text{if } s=\mathrm{CLS} \\
         \tilde{p}_{i, s}\cdot\hat{y}_{i, \mathrm{CLS}}  & \text{if } s \in \{\mathrm{RCA}, \mathrm{LCA}\}\\
         \tilde{p}_{i, s}\cdot\hat{y}_{i, \mathrm{RCA}}   & \text{if } s \in S_{\mathrm{RCA}}\\
         \tilde{p}_{i, s}\cdot\hat{y}_{i, \mathrm{LCA}}  & \text{if } s \in S_{\mathrm{LCA}}\\
    \end{cases}
\end{align}

\subsubsection{Three Level Supervision}
\label{section:training}
We use the binary cross-entropy loss individually on the patient-, artery-, and segment-level targets. We then balance the influence of each level of supervision using three loss coefficients, $P$, $A$, and $S$, corresponding to the patient-, artery-, and segment-levels respectively, such that $P+A+S=1$.
We train for 100 epochs using the AdamW optimizer and cosine-annealing learning rate scheduling.

\subsubsection{Multi-frame Setup}
\label{section:multiframe}
To capture the changes in the artery flow introduced by the temporal dynamics, we design the SegmentMIL to be able to interpret multiple frames from each view. We do so by treating the $K$ frames from a view as individual inputs, increasing the number of input frames from $V_{i}$ to $K \times V_{i}$, $V_{i}$ being the number of views of the patient. To model the temporal ordering within the frames from a view, we introduce temporal embeddings, $\mathbb{R}^{K\times D}$ learned vectors, optimized during the training. Each frame gets assigned one of the temporal embeddings, encoding the relative ordering within the view, thus modeling temporal dynamics while still using image-based encoders.


\begin{table*}[htbp] \centering
 \caption{Comparison of the different variants of the SegmentMIL model against MaxMIL and AttnMIL (classical MIL approaches), and SteDet2Cls (view-level model trained for stenosis detection, used as a classifier). Performance is evaluated at the patient-, view-, artery-, and segment-level on both the internal and CADICA datasets.
 For our SegmentMIL, we evaluate single-level supervision variants: patient (P), artery (A), and segment (S), as well as single and multiple frames per view (FpV) settings, using the optimal configuration of three-level supervision (PAS), with loss coefficients:  $P = 0.4, A = 0.4, S = 0.2$. We report the median AUC with the corresponding 95\% confidence intervals obtained from bootstrapping with 1,000 resampling steps. The overall best model is shown in \textbf{bold}, while the second-best results are \underline{underlined}. When multiple models are underlined, their differences are not statistically significant according to Welch’s t-test ($p < 0.05$). 
 }
\label{tab:main}
    \begin{tabular}{cccccccccc}
    \toprule
    \multicolumn{3}{c}{}& \multicolumn{5}{c}{\textbf{Internal Test Set}}& \multicolumn{2}{c}{\textbf{CADICA}}\\
          \multicolumn{3}{c}{}& \multicolumn{1}{c}{\textbf{Patient (P)}}&\multicolumn{2}{c}{\textbf{Artery (A)}}&\textbf{Segment (S)}&\textbf{View (V)}&\multicolumn{1}{c}{\textbf{Patient (P)}}&\multicolumn{1}{c}{\textbf{View (V)}}\\
          \cmidrule(lr){4-8}
          \cmidrule(lr){9-10}
         \textbf{Method}&\textbf{Supervision}&\textbf{FpV}&\textbf{AUC}&\textbf{RCA AUC}&\textbf{LCA AUC} & \textbf{AUC}& \textbf{AUC}& \textbf{AUC}& \textbf{AUC}\\

        \midrule
             SteDet2Cls*&V&\multirow{5}{*}{1}&\cint{0.639} {0.636, 0.641}&-&-&-&\cint{0.659} {0.657, 0.660}&\cint{0.643} {0.636, 0.648}&\cint{0.631} {0.629, 0.633}\\
         MaxMIL&\multirow{3}{*}{P}&&\cint{0.711} {0.707, 0.711}&-&-&-&\cint{0.683} {0.681, 0.684}&\cint{0.831} {0.820, 0.828}&\cint{0.700} {0.698, 0.702}\\
         AttnMIL~\cite{mil_base}&&&\cint{0.785} {0.781, 0.785}&-&-&-&\cint{0.677} {0.675, 0.678}&\cint{0.785} {0.776, 0.785}&\cint{0.663} {0.661, 0.664}\\
            \cmidrule{1-10}
         \multirow{8}{*}{\rotatebox{90}{\textbf{Segment MIL (Ours)}}}&P&\multirow{5}{*}{1}
         &\cint{\underline{0.832}} {0.828, 0.831}&\cint{\textbf{0.812}} {0.809, 0.813}&\cint{0.732} {0.730, 0.734}&\cint{0.764} {0.763, 0.766}&\cint{\underline{0.693}} {0.690, 0.693}&\cint{0.821} {0.812, 0.820}&
         \cint{\textbf{0.771}} {0.768, 0.772}\\
         &A&&\cint{0.794} {0.791, 0.795}&\cint{0.770} {0.765, 0.770}&\cint{0.776} {0.772, 0.776}&\cint{0.770} {0.767, 0.770}&\cint{0.647} {0.645, 0.648}&\cint{0.815} {0.807, 0.816}&\cint{0.757} {0.755, 0.758}\\
         &S&&\cint{0.783} {0.780, 0.784}&\cint{0.777} {0.774, 0.779}&\cint{0.774} {0.772, 0.776}&\cint{0.762} {0.760, 0.763}&\cint{0.646} {0.644, 0.647}&\cint{0.767} {0.756, 0.766}&\cint{0.636} {0.635, 0.639}\\
         \cmidrule{2-10}
         &\multirow{3}{*}{PAS}&1&\cint{\underline{0.829}} {0.827, 0.830}&\cint{\underline{0.810}} {0.807, 0.812}&\cint{\underline{0.778}} {0.775, 0.780}&\cint{\underline{0.790}} {0.788, 0.791}&\cint{\textbf{0.699}} {0.696, 0.699}&\cint{\underline{0.854}} {0.843, 0.850}&\cint{\underline{0.763}} {0.761, 0.764}\\
         
         &&3&\cint{\textbf{0.845}} {0.841, 0.845}&\cint{\underline{0.809}} {0.806, 0.810}&\cint{\textbf{0.779}} {0.776, 0.780}&\cint{\textbf{0.799}} {0.797, 0.800}&\cint{\underline{0.693}} {0.691, 0.694}&\cint{\textbf{0.878}}{0.869, 0.876}&\cint{0.756} {0.754, 0.757}\\

         \bottomrule
        
         \end{tabular} \caption*{\raggedright\hspace{0.5cm}\footnotesize{* Trained for stenosis detection, evaluated as classifier.}\hfill}

\end{table*}
\subsection{Experimental Setup}
\subsubsection{Evaluation Setup}
We evaluate the models on the internal (\cref{section:clinical_dataset}) and CADICA (\cref{section:cadica_dataset}) datasets.
For each evaluation level (patient, artery, or segment), we report the AUC score. On the artery-level, we distinguish between RCA and LCA, reporting individual performances. On the segment-level, we report the macro average of the AUCs achieved for the $S_{major}$ segments. 
 As the training objective of the SegmentMIL is a weighted sum over the three-level supervision, we conduct experiments to evaluate how different loss coefficients influence performance at each supervision level. Based on these experiments, we identify the optimal loss coefficients that maximizes patient-level AUC, while maintaining a balanced trade-off between artery-, segment-, and view-level performance.
We further investigate the effect of utilizing multiple frames per view by evaluating several frames per view settings and frame-sampling strategies, centered around the key frame. 
 We also do an ablation study across different input image resolutions, encoder backbones, and feature encoding levels, distinguishing between global and patch-level representations. 
Lastly, we use the patch-level attention weights of the SegmentMIL model to identify the input regions that the model attends to. The attention weights are used to generate zero-shot segmentation masks of the arteries, which are evaluated against ground-truth annotations from the ARCADE artery segmentation dataset~\cite{ARCADE}. 

\subsubsection{Baselines}
Current approaches for stenosis diagnosis focus on view-level predictions \cite{rca_stenosis,cadica, stenosis_3_view_quantification}, with the work in \cite{stenosis_image_and_patient_level_cls} being the only approach performing patient-level evaluation. However, as this approach still relies on view-level annotations and model weights and training data are not publicly available, we can not reproduce the model. Instead, we implement three baseline methods: \emph{SteDet2Cls}, \emph{MaxMIL}, and \emph{AttnMIL}.

SteDet2Cls is a YOLO-based\cite{yolo} object detection model, trained on the ARCADE stenosis detection dataset, and in this study, we use it as a view-level binary classifier. We train SteDet2Cls using view-level supervision and infer patient-level prediction via max-pooling.
We also introduce MaxMIL~\cite{max_mil}, an instance-level pooling-based MIL, which aggregates view-level predictions using max-pooling~\cite{max_vs_noisy_or}, and, same as SegmentMIL, is trained on patient-level annotations. 
Last, we introduce AttnMIL~\cite{mil_base}, a classical attention-based MIL approach, widely used in the histopathology domain~\cite{rethinking_attention_based_mil,clam,xmil}. 
In this approach we aggregate the features from the different views  at the embedding-level using attention and train it on the same supervision as MaxMIL and SegmentMIL.\label{section:experiments}

\section{Results}

\subsection{Main Results and Key Findings}
 In \cref{tab:main}, we present a comparison between the SteDet2Cls, MaxMIL, and AttnMIL baselines as well as various configurations of our proposed SegmentMIL. For SegmentMIL, we evaluate models trained with single-level supervision, either patient (P), artery (A), or segment (S), as well as with three-level supervision (PAS) using the best performing combination of loss coefficients ($P = 0.4, A = 0.4, S = 0.2$), and the best-performing multi-frame configuration.
The comparison is conducted on both the internal and CADICA datasets.  
For each configuration, we report the median AUC, along with the 95\% confidence interval, estimated using bootstrapping with 1,000 resampling steps. Statistical significance is assessed using a Welch’s t-test with a significance threshold of $p < 0.05$. The patient-level AUC is our primary performance indicator.

\textbf{\noindent{SegmentMIL demonstrates best performance across both internal and external evaluations, outperforming classical MIL and view-level approaches.}} The best-performing configuration (multi-frame setting with three-level supervision) achieves patient-level AUCs of 0.845 and 0.878 on the internal and CADICA datasets, significantly outperforming other SegmentMIL configurations and baselines. Both single- and multi-frame PAS SegmentMIL outperform MaxMIL and AttnMIL in all evaluation categories on both datasets, with the multi-frame setting outperforming MaxMIL by 13\% AUC, and AttnMIL by 6\% AUC on the patient-level on the internal test set. The performance gap is even larger when compared to the SteDet2Cls model, exceeding 20\% patient-level AUC. The big performance gap between SegmentMIL and SteDet2Cls is a result of the overprediction of positive cases by the SteDet2Cls model (large recall and low precision in both datasets), a direct result of the object detection training, where the model has not seen any stenosis-free samples.
The single-frame SegmentMIL model trained with only patient-level supervision exhibits similar performance to the three-level supervised models, outperforming MaxMIL, AttnMIL, and SteDet2Cls at the patient-level. The only exception is the evaluation of MaxMIL on the CADICA dataset, where MaxMIL is comparable to the single-frame SegmentMIL trained with patient-level supervision. 
MaxMIL has strong performance in this setting because in CADICA patients with more views are more likely to have stenosis (\cref{fig:data:b}). Since MaxMIL’s max-pooling tends to increase the predicted probability with more views, it leverages this dataset bias.


\textbf{\noindent{Three-level supervision consistently outperforms single-level supervision across all evaluation levels.}}
When comparing the single-frame models, the results indicate that the SegmentMIL trained with three-level supervision (PAS) achieves among the top-two performance across all evaluation settings, ranking as best in four, and second-best in three settings. The only single-frame model competitive to the three-level SegmentMIL is the SegmentMIL trained on patient-level supervision, which, even though it achieves best patient-level performance on the internal test set, the difference to the three-level SegmentMIL is not statistically significant. Further, the three-level SegmentMIL is the only model achieving consistently strong performance across all evaluation levels, confirming the benefit of the three-level supervision.

\noindent\textbf{Introducing the temporal dimension through multiple frames per view improves performance.}
Although modeled using an image-based encoder, the usage of multiple frames per angiography view yields performance improvements in all evaluation levels apart from the view-level, with the patient-level improvements being 1.6\% and 2\% AUC on the internal and CADICA dataset. Such performance aligns with clinical practice, where cardiologists assess the temporal dynamics of contrast flow through the coronary arteries to identify stenosis. The observed performance gain highlights the value of incorporating temporal context and suggests that further improvements could be achieved by employing video-based encoders that explicitly model temporal dependencies.

\noindent\textbf{SegmentMIL achieves strong view-level performance despite being trained with patient-level annotations.}
Although primarily designed for patient-level stenosis classification, SegmentMIL also demonstrates strong performance at the view-level. The  multi-frame three-level SegmentMIL (best configuration) surpasses the SteDet2Cls model, trained explicitly for view-level stenosis detection, by 3\% and 12\% AUC on both internal and CADICA datasets. Moreover, it outperforms the MaxMIL model, trained for instance-level MIL over view-level predictions. These results indicate that, despite patient-level supervision, our SegmentMIL effectively learns a view-level representation.

\subsection{Ablations}

\subsubsection{Three Level Supervision
Ablation}

\begin{figure}[!t]
  \centering
  \begin{subfigure}[b]{\linewidth}
  \centering
     \includegraphics[width=\linewidth]{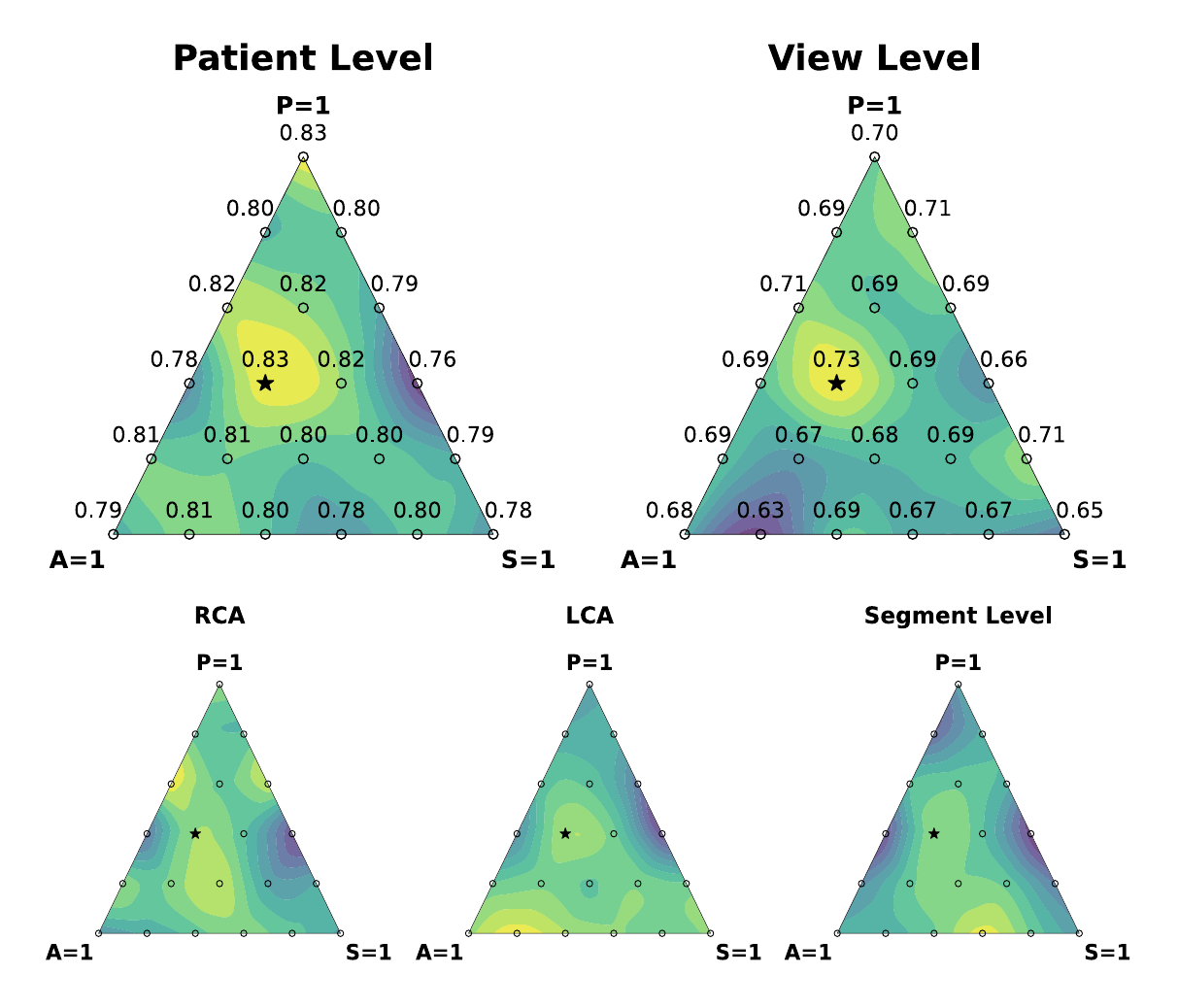}
  \caption{Internal Test Set} 
  \end{subfigure}

 \begin{subfigure}[b]{\linewidth}
  \centering
     \includegraphics[width=\linewidth]{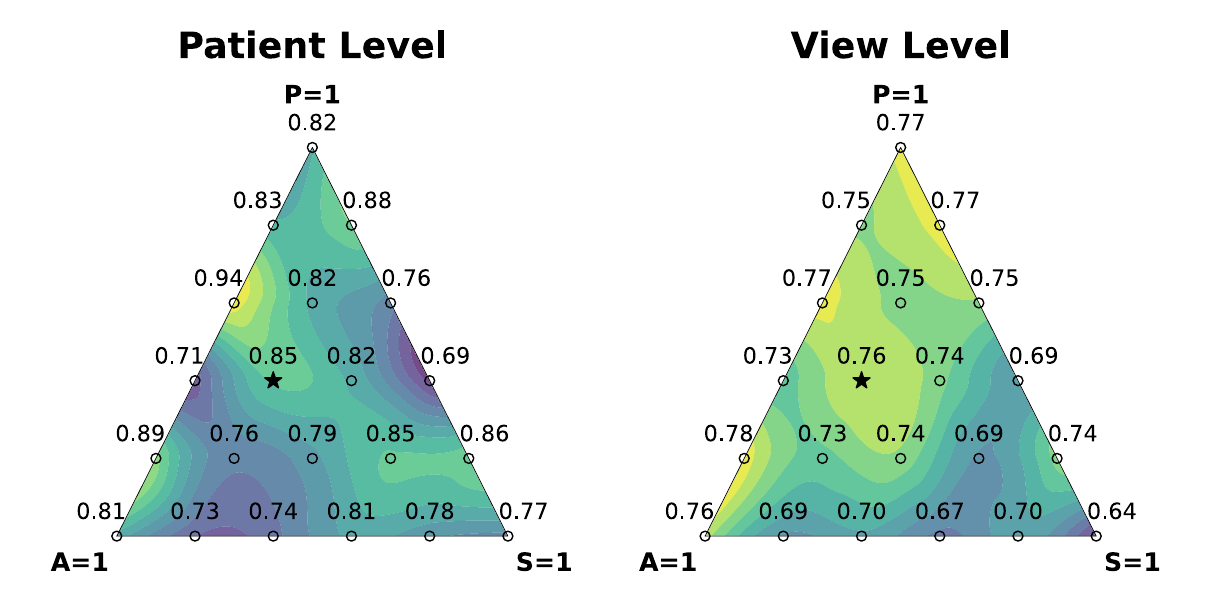}
  \caption{CADICA Test Set}
     \end{subfigure}
 \caption{Influence of the loss coefficients across the different evaluation levels and datasets, shown as simplex plots. For both datasets, we show AUC at the patient- and view-level. For the internal dataset, we also show the AUC at RCA, LCA, and segment (reported as macro average) levels. The loss weights range from 0 to 1, sampled in steps of 0.2, with intermediate values obtained by interpolation. Each vertex of the plot represents supervision from a single level, and its influence decreases as we move away from that vertex and reaches zero at the opposite edge. The optimal loss coefficients configuration is marked by a $\bigstar$. We see that, for both datasets, the patient- and view-level exhibit similar trends, also reflected in the RCA. In contrast, the LCA and segment-level plots have distinct distributions, indicating differing dynamics across these levels.
  }
  \label{Fig:simplex}
  \end{figure}

We investigate the effect of the loss coefficients on the performance at each evaluation level (patient-, artery-, and segment-levels), via simplex plots, shown in \cref{Fig:simplex}. As the coefficients sum to one (\cref{section:training}), we explore all coefficient combinations within the $[0,1]$ range, categorized in steps of 0.2, and obtain the intermediate results via interpolation. The three vertices of a plot correspond to a single-level supervision configuration. The value of the loss coefficient decreases the further we are from its vertex, reaching zero along the edge opposite to it.
The plots reveal similar performance dynamics between the patient- and view-level evaluations across both datasets, where the best performing setups are obtained when focusing on both the patient- and artery-level supervision, with minimal segment-level supervision. Similar dynamics are also obtained for the RCA. In contrast, the LCA and segment-level plots follow distinct trends, where the best results are achieved when the supervision is focused on the artery- and segment-level, respectively, with minimal patient-level supervision. This behavior suggests that, due to the increased complexity of segments and finer granularity of LCA  compared to RCA, these tasks benefit from more specific supervision signals.

\subsubsection{Encoder Architecture Ablation} 
We compare different SegmentMIL's encoder architectures (\cref{tab:ablation}), evaluating the patient-level AUC on the internal test set using ResNet-50 and ViT-S (patch size 14) backbones across two input resolutions and encoding levels.
The results show that increasing the input resolution to $518\times518$ improves AUC for the patch-based encodings of both backbones, highlighting the importance of fine-grained details for stenosis classification.
Furthermore, the encoding level has a strong impact on the ViT-based models: the patch-level encoding achieves 12\% higher AUC compared to the global encoding at high resolution, caused by the richer information within the patch embeddings. In contrast, the ResNet backbone shows limited sensitivity to the encoding level.
Overall, the best performance is obtained with a patch-level ViT-S encoder trained on $518\times518$ input resolution, which is adopted for all experiments.

\begin{table}[t]\centering
\caption{
Ablation study on SegmentMIL's encoder, evaluated on patient-level AUC on the internal test set. We analyze the influence of input resolution, backbone model, and encoding level. The ViT outperforms the ResNet, achieving the best performance with high-resolution input and patch-level encoding.}
\scriptsize
    \begin{tabular}{ccccc}
    \toprule
        \textbf{Train Level}&\textbf{Resolution}&\textbf{Backbone}&\textbf{Encode Level}&\textbf{AUC}\\
         \midrule
         \multirow{9}{*}{P}&\multirow{4}{*}{224}&\multirow{2}{*}{ResNet50}&Global&0.797\\
         &&&Patch&0.780\\
         \cmidrule{3-5}
         &&\multirow{2}{*}{ViT-S/14}&Global&0.730\\
         &&&Patch&\underline{0.804}\\
\cmidrule{2-5}
         &\multirow{4}{*}{518}&\multirow{2}{*}{ResNet50}&Global&0.781\\
         &&&Patch&0.795\\
         \cmidrule{3-5}
         &&\multirow{2}{*}{ViT-S/14}&Global&0.711\\
         &&&Patch&\textbf{0.832}\\
         \bottomrule
    \end{tabular}
    
    \label{tab:ablation}
\end{table}

\begin{figure}[t!]
  \centering
     \includegraphics[width=\linewidth]{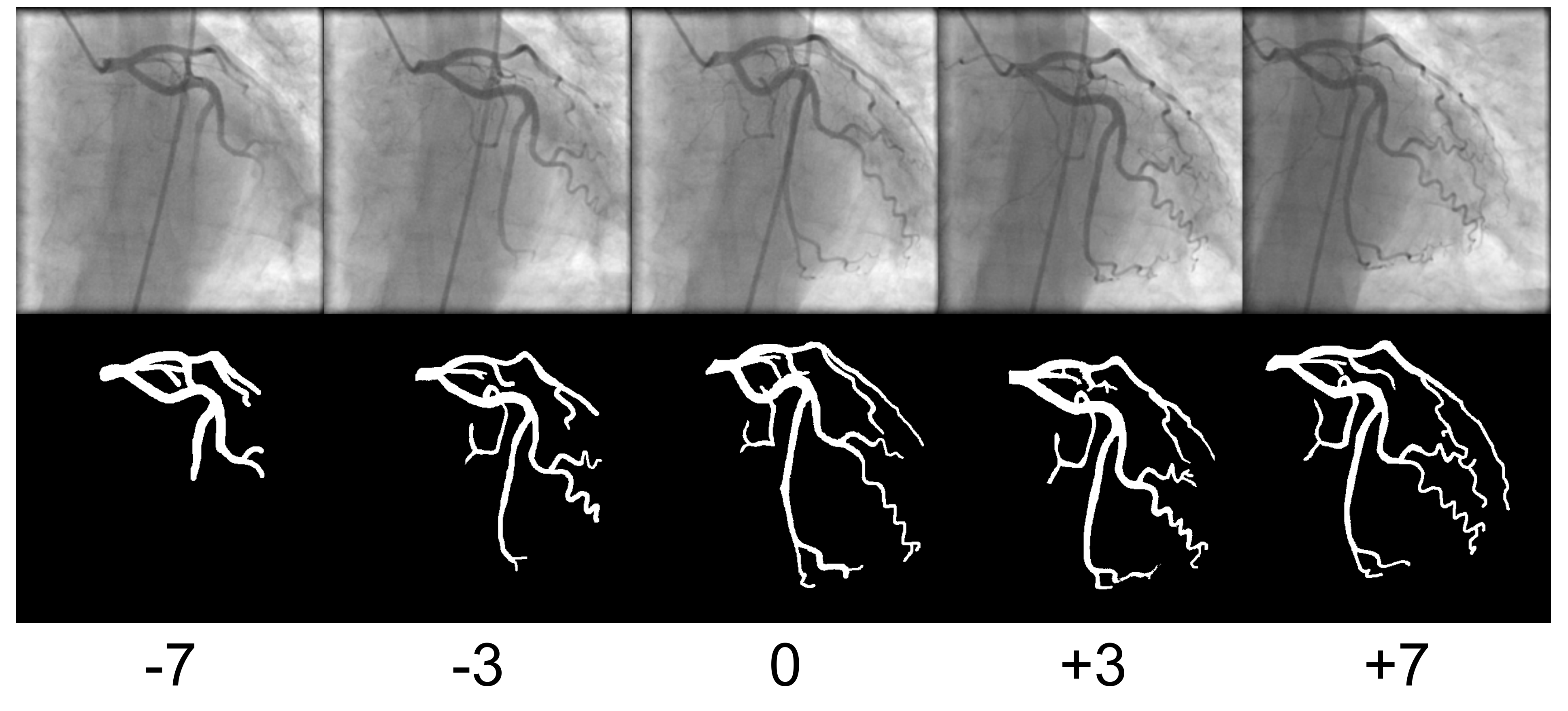}
\caption{Changes in arteries across the temporal dimension within a single view. The segmentation masks are manually annotated to highlight the arteries visible across $\pm$7 frames (corresponding to 2 seconds) centered around the key frame.}\label{fig:multiframe} 
  \end{figure}

\subsubsection{Multi-frame Ablations}
By focusing on a single frame per view, the model neglects the temporal flow dynamics in angiography sequences. In \cref{fig:multiframe} we show the change in contrast and artery visibility across $\pm$7 frames around a key frame, corresponding to two seconds of a video. To assess the benefit of the temporal dimension, we do an ablation over the number of frames used per view (using between two and five), exploring multiple frame selection strategies centered around the key frame. The resulting patient-level AUC values on the internal test set are reported in \cref{tab:fpv_ablation}. The results show that even including a single additional frame yields a measurable performance gain, peaking at 2\% AUC when using three frames per view. Among the tested frame selection strategies, the best performance is achieved when using frames closest to the key frame, i.e., frames that are most visually similar to the key frame. Such configurations introduce minimal temporal change while still achieving the best performance, suggesting that the performance improvement does not directly follow the amount of temporal change. This behavior may be attributed either to our strategy for encoding temporal relations or to the noise introduced by including dissimilar frames.

  \begin{table}[t!] \centering
\caption{Ablation of the number of frames per view, evaluated on the patient-level internal test set. For each number of frames, we assess multiple frame distributions centered around the key frame (denoted as 0). The best configuration within each group is shown in \textbf{bold}, and the second-best is \underline{underlined}. 
The best setting uses three frames per view, with frames distributed closely around the key frame.}
\scriptsize
    \begin{tabular}{cr@{\hspace{3pt}}c@{\hspace{3pt}}lc}
    \toprule
        \textbf{Frames per View}&\multicolumn{3}{c}{\textbf{Frames Distribution}}&\textbf{AUC}\\
         \midrule
         1&&0& &\textbf{0.829} \\
         \cmidrule{1-5}
            \multirow{4}{*}{2}&-1&0& &\textbf{0.835} \\
         &&0&+1&\textbf{0.835}\\
         &-3&0&&0.823\\
         &&0&+3&\underline{0.826}\\
         \cmidrule{1-5}
         \multirow{4}{*}{3}&-1&0&+1&\textbf{0.845}\\
         &-2 -1&0&&\underline{0.835}\\
         &&0&+1 +2&0.830 \\
         &-3&0&+3&0.823 \\
         \cmidrule{1-5}
         \multirow{4}{*}{4}&-2 -1&0&+1&\underline{0.838}\\
         &-1&0&+1 +2&\textbf{0.840}\\
        &-3 -2 -1&0&&0.832 \\
         &&0&+1 +2 +3&0.833\\
         \cmidrule{1-5}
         \multirow{3}{*}{5}&-2 -1&0&+1 +2&\textbf{0.836} \\
         &-3 -2 -1&0&+1&\underline{0.835} \\
        &-1&0&+1 +2 +3&0.834\\

         \bottomrule
    \end{tabular}
    \label{tab:fpv_ablation}
\end{table}
\subsection{Attention Interpretation}
\begin{figure*}[t!]
  \centering
     \includegraphics[width=\linewidth]{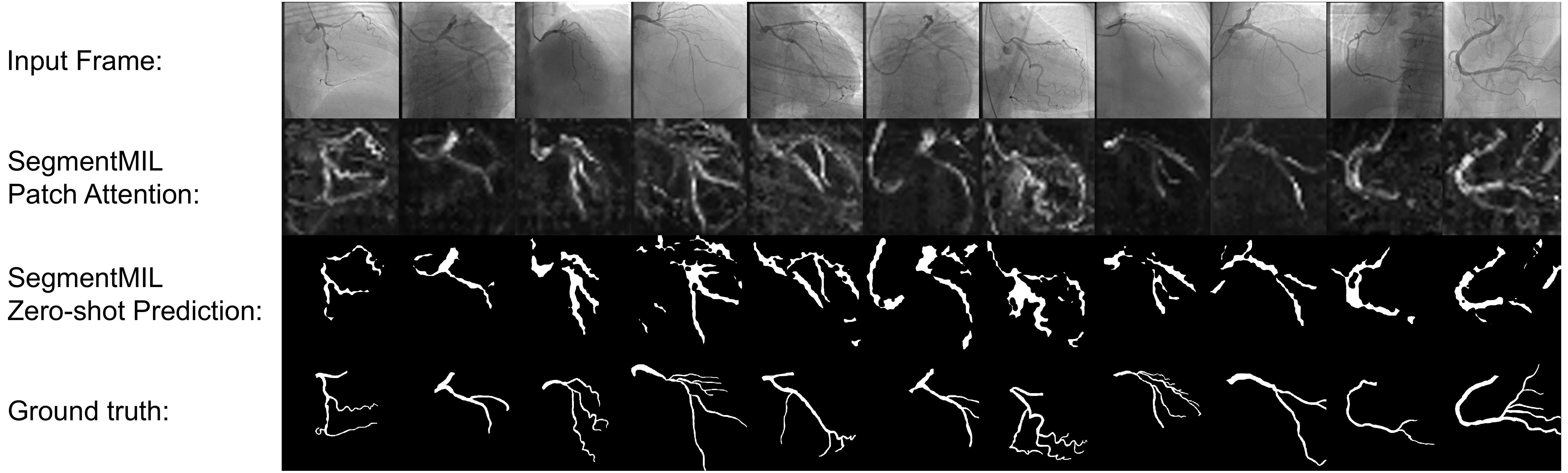}
  \caption{Comparison of zero-shot artery segmentation predictions against manually annotated ground truth from the ARCADE artery segmentation dataset. The predicted masks, shown in the third row, are obtained by binarizing the patch-wise attention weights (second row) produced by our best-performing SegmentMIL model for patient-level stenosis classification. The results demonstrate that the model strongly focuses on the relevant anatomical structures and, while coarse, it even successfully captures smaller artery regions that are present but not labeled in the ground truth annotations.}\label{fig:segmentation} 
  \end{figure*}

%

As described in \cref{section:experiments}, we leverage the patch-level attention weights from SegmentMIL to analyze the image regions to which the models attend the most. From the attention maps, we derive zero-shot artery segmentation masks, which are compared against the ground truth annotations from the ARCADE artery segmentation dataset. Qualitative evaluation of the obtained masks is shown in \cref{fig:segmentation},
which shows that apart from larger artery segments, the obtained masks often also capture smaller parts of arteries that are visible in the input frames but not annotated in the ground truth, suggesting that the model correctly attends to artery structures at different scales. We also observe that SegmentMIL performs particularly well on RCA cases (illustrated in the last two examples), which are structurally less granular. In these cases, the model’s attention maps almost perfectly align with the arteries. Such behavior indicates that SegmentMIL effectively focuses on disease-relevant regions, without any localization supervision, thereby enhancing the trustworthiness of its predictions.

\section{Discussion and Conclusion}
\subsection{Discussion}
Our SegmentMIL diagnoses patient-level stenosis from multiple views, achieving AUCs of 0.845 and 0.878 on internal and external evaluation, respectively. It outperforms the classical MIL approaches as well as the view-level methods introduced in this study  (\cref{tab:main}). SegmentMIL achieves comparable performance to the work in ~\cite{stenosis_image_and_patient_level_cls}, which is the only work that does artery- and patient-level evaluation, reporting AUCs of 0.89, 0.84, and 0.86 for RCA, LCA, and patient-level, respectively. What distinguishes the evaluation of the work of \cite{stenosis_image_and_patient_level_cls} and our SegmentMIL is that they evaluate using a stenosis threshold of 25\%, and a fixed number of views captured from predefined angulations (7 views in total, 3 for RCA, and 4 for LCA), 
while SegmentMIL is evaluated on real-world clinical data with a stenosis threshold of 70\%, indicating severe clinically relevant stenosis. Furthermore, while we use a single model capable of handling any number and angulation of views, in ~\cite{stenosis_image_and_patient_level_cls} they train different models for the specific angulations. Last, similar to other methods for stenosis diagnosis\cite{stenosis_detection_extra_1,stenosis_detection_extra_2,stenosis_detection_extra_spatio_temporal,rca_stenosis}, the work in \cite{stenosis_image_and_patient_level_cls} relies on view-level annotations, obtained through an expensive and time-consuming manual labeling. This however, is not the case for SegmentMIL, which reuses patient-level annotations already present in hospital systems, and does not require any manual labeling. Furthermore, SegmentMIL is the only approach that models stenosis classification primarily as a patient-level task, capable of training using multiple frames from a view. It is also the only model providing patient-, artery-, and segment-level predictions, which adds transparency and helps in clinical decision making. 
Moreover, the analysis of the patch attention weights shows that the SegmentMIL strongly attends to the visible arteries in the angiography view, which are the exact artifacts relevant for stenosis diagnosis, increasing the trustworthiness of the  SegmentMIL's performance. 
\subsection{Limitations}
The main limitation of this study lies in the use of image-based encoders for both single and multi-frame settings. Although performance improvements are achieved by modeling temporal relations through learned temporal embeddings (\cref{section:multiframe}), this approach only partially captures the dynamics present in angiography sequences. However, this indicates a promising direction for future work, where employing video-based models could more effectively leverage the temporal information present in the data.
Furthermore, we train our SegmentMIL using only a subset of coronary artery segments. While such a setup focuses on larger segments, more relevant for placing a stent, it also means the model does not learn to recognize stenosis that may arise in other parts of the angiogram.
Lastly, our study relies solely on angiography imaging, without incorporating additional patient context such as medical history, clinical symptoms, or complementary examination results. These factors are available to cardiologists during diagnosis, and could provide valuable information and further enhance model performance and clinical relevance.
\subsection{Conclusion}

Current deep-learning methods for stenosis diagnosis focus on view-level models, which rely on manual annotations and do not consider the multi-view nature of angiography data. We overcome this limitation by using MIL, allowing us to train stenosis classification models using reliable annotations already available in hospital systems. 
This study is the first work that focuses on patient-level stenosis diagnoses, and the strong performance of our SegmentMIL is proof that MIL can be used on raw hospital annotations. 
As there still exists a gap in performance between the SegmentMIL and trained cardiologists, the future work will focus on addressing the limitations of this study with a goal of producing even stronger models whose fast and reliable assessments will assist cardiologists in time-constrained environments.

\bibliographystyle{splncs04}

\bibliography{sections/references}

\end{document}